\documentclass{article}

\usepackage{amsmath}

\DeclareMathOperator*{\argmin}{\arg\!\min}

\usepackage{todonotes}
\usepackage{multirow}
\usepackage{tikz}
\usepackage{pgfplots}
\pgfplotsset{
    table/search path={results},
    compat=1.14
}
\usepackage{subcaption}

\newtheorem{definition}{Definition}[section]
\newtheorem{proposition}{Proposition}[section]

    \PassOptionsToPackage{numbers, compress}{natbib}
\usepackage[preprint]{neurips_2020}

\usepackage[utf8]{inputenc} %
\usepackage[T1]{fontenc}    %
\usepackage{hyperref}       %
\usepackage{url}            %
\usepackage{booktabs}       %
\usepackage{amsfonts}       %
\usepackage{nicefrac}       %
\usepackage{microtype}      %

\title{Manifold Regularization for \\ Locally Stable Deep Neural Networks}

\author{%
  Charles Jin \\
  CSAIL\\
  MIT\\
  Cambridge, MA 02139 \\
  \texttt{ccj@csail.mit.edu} \\
  \And
  Martin Rinard \\
  CSAIL\\
  MIT\\
  Cambridge, MA 02139 \\
  \texttt{rinard@csail.mit.edu} \\
}

\begin{document}

\maketitle

\begin{abstract}
  We apply concepts from manifold regularization to develop new regularization techniques for training locally stable deep neural networks. Our regularizers are based on a sparsification of the graph Laplacian which holds with high probability when the data is sparse in high dimensions, as is common in deep learning. Empirically, our networks exhibit stability in a diverse set of perturbation models, including $\ell_2$, $\ell_\infty$, and Wasserstein-based perturbations; in particular, we achieve 40\% adversarial accuracy on CIFAR-10 against an adaptive PGD attack using $\ell_\infty$ perturbations of size $\epsilon = 8/255$, and state-of-the-art verified accuracy of 21\% in the same perturbation model. Furthermore, our techniques are efficient, incurring overhead on par with two additional parallel forward passes through the network.
\end{abstract}

\section{Introduction}

Recent results in deep learning highlight the remarkable performance deep neural networks can achieve on tasks using data from the natural world, such as images, video, and audio. Though such data inhabits an input space of high dimensionality, the physical processes which generate the data often manifest significant biases, causing realistic inputs to be sparse in the input space.

One way of capturing this intuition is the manifold assumption, which states that input data is not drawn uniformly from the input space, but rather supported on some smooth submanifold(s) of much lower dimension. Starting with the work of \citet{belkin2006manifold}, this formulation has been studied extensively in the setting of semi-supervised kernel and regression methods, where algorithms exploit the unlabelled data points to learn functions which are smooth on the input manifold \citep{geng2012ensemble, goldberg2008online, niyogi2013manifold, sindhwani2005linear, tsang2007large, xu2010discriminative}. Comparatively, such techniques have seen less use in the context of deep neural networks, owing in part to the ability of such models to generalize from relatively sparse data \citep{zhang2016understanding}.

\paragraph{Contributions} We apply concepts from manifold regularization to train locally stable deep neural networks. In light of recent results showing that neural networks suffer widely from adversarial inputs \citep{szegedy2013intriguing}, our goal is to learn a function which does not vary much in the neighborhoods of natural inputs, even when the network classifies incorrectly. We show that this definition of local stability has a natural interpretation in the context of manifold regularization, and propose an efficient regularizer based on an approximation of the graph Laplacian when data is sparse in high dimensions. Crucially, our regularizer exploits the continuous piecewise linear nature of ReLU networks to learn a function which is smooth over the data manifold in not only its outputs but also its decision boundaries.

We evaluate our approach by training neural networks with our regularizers for the task of image classification on CIFAR-10 \citep{krizhevsky2009learning}. Empirically, our networks exhibit robustness against a variety of adversarial models implementing $\ell_2$, $\ell_\infty$, and Wasserstein-based attacks. We also achieve state-of-the-art \textit{verified} robust accuracy under $\ell_\infty$ of size $\epsilon = 8/255$. Furthermore, our regularizers are cheap: we simply evaluate the network at two additional random points for each training sample, so the total computational cost is on par with three parallel forward passes through the network. Our techniques thus present a novel, regularization-only approach to learning robust neural networks, which achieves performance comparable to existing defenses while also being an order of magnitude more efficient.

\section{Background}

\paragraph{Manifold regularization}

The manifold assumption states that input data is not drawn uniformly from the input domain $\mathcal{X}$, also know as the \textit{ambient space}, but rather is supported on a submanifold $\mathcal{M} \subset \mathcal{X}$, called the \textit{intrinsic space}. There is thus a distinction between regularizing on the ambient space, where the learned function is smooth with respect to the entire input domain (e.g., Tikhonov regularization \citep{phillips1962technique, tikhonov2013numerical}), and regularizing over the intrinsic space, which uses the geometry of the input submanifold to determine the regularization norm. 

A common form of manifold regularization assumes the gradient of the learned function $\nabla_\mathcal{M} f(x)$ should be small wherever the probability of drawing a sample is large. We will refer to such functions as ``smooth''. Let $\mu$ be a probability measure with support $\mathcal{M}$. This idea leads to the following intrinsic regularizer:
\begin{equation}
\label{eq:reg_integral}
    ||f||^2_I := \int_\mathcal{M} ||\nabla_\mathcal{M} f(x)||^2 d\mu(x)
\end{equation}
In general, we cannot compute this integral because $\mathcal{M}$ is not known, so \citet{belkin2006manifold} propose the following discrete approximation that converges to the integral as the number of samples grows \citep{BELKIN20081289}:
\begin{equation}
\label{eq:reg_discrete}
    ||f||^2_I \approx \frac{1}{N^2} \sum_{i, j = 1}^{N} (f(x_i) - f(x_j))^2 L_{i,j}
\end{equation}
Here, the $x_1, ..., x_N$ are samples drawn, by assumption, from the input manifold $\mathcal{M}$ according to $\mu$, and $L$ is a matrix of weights measuring the similarity between samples. The idea is to approximate the continuous input manifold using a discrete graph, where the vertices are samples, the edge weights are distances between points, and the Laplacian matrix $L$ encodes the structure of this graph. A common choice of weights is an isotropic Gaussian kernel $L_{i,j} := \exp(-||x_i - x_j||^2 / s)$. To increase the sparsity of the Laplacian, weights are often truncated to the $k$-nearest neighbors or within some $\epsilon$-ball. Note that the Laplacian can also be interpreted as a discrete matrix operator, which converges under certain conditions to the continuous Laplace operator \cite{BELKIN20081289}.

\paragraph{ReLU networks}
Our development focuses on a standard architecture for deep neural networks: fully-connected feedforward networks with ReLU activations. In general, we can write the function represented by such a network with $n$ layers and parameters $\theta = \{A_i, b_i\}_{i = 1, ..., n-1}$ as 
\begin{align}
    z_0 &= x                                                                \\
    \hat{z}_i &= A_i \cdot z_{i-1} + b_i    &\text{for $i = 1, ..., n-1$}   \\
    \label{eq:relu_postact}
    z_i &= \sigma(\hat{z}_i)                &\text{for $i = 1, ..., n-2$}   \\
    f(x; \theta) &= \hat{z}_{n-1}
\end{align}
where the $A_i$ are the weight matrices and the $b_i$ are the bias vectors. We call the $z_i$ ``hidden activations'', or more simply, activations, and the $\hat{z}_i$ ``pre-activations''.

In this work, we consider ReLU networks, where the non-linear activation function $\sigma(\cdot)$ in (\ref{eq:relu_postact}) is the Rectified Linear Unit (ReLU)
\begin{align*}
    z_i &= \sigma(\hat{z}_i) := \max(0, \hat{z}_i)
\end{align*}
It is clear from this description that ReLU networks are a family of continuous piecewise linear functions. We denote the linear function induced by an input $x$ as $f_x(\cdot; \theta)$, i.e., the analytic extension of the local linear component about $x$ over the input domain.

\paragraph{Adversarial robustness}
One common measure of robustness for neural networks is against a norm-bounded adversary. In this model, the adversary is given an input budget $\epsilon$ over a norm $||\cdot||_{in}$, and asked to produce an output perturbation $\delta$ over a norm $|\cdot|_{out}$. A point $x'$ is an $\epsilon$-$\delta$ adversarial example for an input pair $(x, y)$ if
\begin{align*}
    ||x' - x||_{in} &\le \epsilon \\
    |f(x'; \theta) - y|_{out} &\ge \delta
\end{align*}
When the specific norm is either unimportant or clear from context, we also write the first condition as $x' \in N_\epsilon(x)$, where $N_\epsilon(x)$ refers to the $\epsilon$-ball or neighborhood about $x$. If such an adversarial example does not exist, we say that the network is $\epsilon$-$\delta$ robust at $x$.

Standard examples of $||\cdot||_{in}$ include the $\ell_2$ and $\ell_\infty$ ``norm'', defined for vectors as $||x||_\infty := \max_i |x_i|$. For classification tasks, the adversary is successful if it produces an example in the $\epsilon$-neighborhood of $x$ which causes the network to misclassify. In this case, we drop $\delta$ and say that the network is $\epsilon$-robust at $x$. Note that if $f(x; \theta)$ is already incorrect, then $x$ suffices as an adversarial example. 

\section{Related work}

Manifold regularization was first introduced by \citet{belkin2006manifold} in the context of semi-supervised learning, where the goal was to leverage unlabeled samples to learn a function which behaves well (e.g., is smooth, or has low complexity) over the data manifold. The use of manifold regularization for deep neural networks has been explored in several contexts. \citet{tomar2014manifold, tomar2016graph} consider the use of manifold regularization for speech recognition, where they argue that human vocal cords parameterize a low dimensional input manifold. \citet{hu2018robust} note that object tracking datasets contain large amounts of unlabelled data and propose to use manifold regularization to learn from the unlabelled data. \citet{zhu2018ldmnet} apply manifold regularization so that the concatenation of input data and output features comes from a low dimensional manifold, and find features transfer better to different modalities. \citet{lee2015manifold} combine manifold regularization with adversarial training and show improvements in standard test accuracy. Our approach is to use manifold regularization to induce stability separately from accuracy. We note that a similar decomposition between accuracy and stability forms the basis for the TRADES algorithm \citep{zhang2019theoretically}, though the training procedure ultimately relies on adversarial training. \citet{hein2017formal} propose a conceptually similar regularizer to minimize the difference between logits and show improved $\ell_2$ certified robustness. Finally, \citet{tsang2007large} derive a similar sparsified approximation of the Laplacian in the semi-supervised setting when the samples are dense; our techniques and analysis differ primarily because we induce sparsity via resampling as we cannot rely on the inherent properties of our data.

Adversarial examples were introduced by \citet{szegedy2013intriguing}, who found that naively trained neural networks suffer almost complete degradation of performance on natural images under slight perturbations which are imperceptible to humans. A standard class of defense is \textit{adversarial training}, which is characterized by training on adversarially generated input points \citep{dong2017boosting, goodfellow2014explaining, maini2019adversarial, pang2019rethinking, shafahi2019adversarial, wong2019wasserstein, wong2020fast, xie2018feature, zhang2019theoretically}. In particular, the basic Project Gradient Descent (PGD) attack \citep{aleks2017deep} is widely considered to be an empirically sound algorithm for both training and evaluation of robust models. However, such training methods rely on solving an inner optimization via an iterative method, effectively increasing the number of epochs by a multiplicative factor (e.g., an overhead of 5--10x for standard PGD \citep{shafahi2019adversarial}, though recent work suggests early stopping can help \citep{wong2020fast}). In comparison, our techniques do not rely on the construction of any worst-case inputs and are more efficient than standard adversarial training.

Another approach is to train models which are \textit{provably robust}. One method is to use an exact verification method, such as an MILP solver, to prove that the network is robust on given inputs \citep{tjeng2017evaluating}. In particular, \citet{xiao2019training} use a similar loss based on ReLU pre-activations to learn stable ReLUs for efficient verification, but rely on a PGD adversary to train a robust model. Certification methods modify models to work directly with neighborhoods instead of points \citep{dvijotham2018training, gowal2018effectiveness,mirman2018differentiable, wong2018scaling}. In practice, the inference algorithms must overapproximate the neighborhoods to preserve soundness while keeping the representation compact as it passes through the network. This strategy can be interpreted as solving a convex relaxation of the exact verification problem. Though certification thus far has produced better lower bounds, verification as a technique is fully general and can be applied to any model (given sufficient time); recent work also suggests that methods using layerwise convex relaxations may face an inherent barrier to tight verification \citep{NIPS2019_9176}.

\section{Setting}

We reframe the goal of learning functions that are robust using a perspective which decouples stability from accuracy. The key observation is that we would like to train networks that are locally stable around natural inputs, even if the network output is incorrect. This approach contrasts with adversarial training, which attempts to train the network to classify correctly on worst-case adversarial inputs. In particular, recall that a network is $\epsilon$-robust at $x$ if no point in the $\epsilon$-neighborhood of $x$ causes the network to misclassify. We consider the related property of $\epsilon$-stability:
\begin{definition}
    A function $f$ is \emph{$\epsilon$-$\delta$ stable} at an input $x$ if for all $x' \in N_\epsilon(x)$, $|f(x) - f'(x)| \le \delta$. A classifier $f$ is \emph{$\epsilon$-stable} at an input $x$ if for all $x' \in N_\epsilon(x)$, $f(x) = f(x')$.
\end{definition}
This definition is independent of the correct label for $x$; as such, we argue that $\epsilon$-stability is a property of the function with respect to the input manifold and can thus be captured using manifold regularization. For completeness, we state the following connection between robustness and stability:
\begin{proposition}
    A function $f$ is \emph{$\epsilon$-$\delta$ robust} at an input $x$ iff $f$ is $\epsilon$-$\delta$ stable at $x$ and $f(x) = y$. A classifier $f$ is \emph{$\epsilon$-robust} at an input $x$ iff $f$ is $\epsilon$-stable at $x$ and $f$ correctly classifies $x$.
\end{proposition}
We explore the hypothesis that optimizing for stability and accuracy separately will result in a function which is robust as a byproduct.

\section{Manifold regularization for deep neural networks}

Assume that we have a function $f$ such that $||\nabla f(x')|| \le \delta / \epsilon$ for $x' \in N_\epsilon(x)$. Then clearly $f$ is $\epsilon$-$\delta$ stable at input $x$. This suggests that minimizing the norm of the gradient of $f$ might also induce $\epsilon$-stability in $f$. However, rather than minimize the gradient of $f$ everywhere, we note that we only need $\epsilon$-stability around valid input points; thus we arrive at manifold regularization.

Applying the regularization term in Equation \ref{eq:reg_discrete} yields, in the limit, a function which is smooth on the data manifold. Unfortunately, a straightforward approach does not suffice for our goal of learning $\epsilon$-stable deep neural networks. The first problem is that convergence of the discrete approximation requires that the samples be dense over the input manifold; however, this assumption is almost certainly violated in most practical applications, particularly in the deep learning regime. The second problem is that smoothness on the data manifold does not yield $\epsilon$-stability; indeed, smoothness as defined in Equation \ref{eq:reg_integral} is a global characteristic with average-case guarantees, whereas stability is a local property with worst-case behavior. The next two sections are dedicated to these challenges.

\subsection{Resampling for local smoothness}

We write the $\epsilon$-neighborhood of a manifold $\mathcal{M}$ as $\mathcal{M}_\epsilon := \{x : \exists y \in \mathcal{M}, ||x - y|| \le \epsilon\}$. Since $\epsilon$-stability is defined over the $\epsilon$-neighborhood of every input point, we might want a function which is smooth over $\mathcal{M}_\epsilon$ instead of just $\mathcal{M}$, e.g., by slightly perturbing every input point. This strategy does produce samples from the $\epsilon$-neighborhood of the data manifold, however note that the induced measure places less density at the boundaries. Nevertheless, this procedure exhibits several properties we can exploit. The first is that for sufficiently small $\epsilon$, we get nearly the same operator over $\mathcal{M}_\epsilon$ and $\mathcal{M}$, so that smoothness over $\mathcal{M}_\epsilon$ does not sacrifice smoothness on the original data manifold $\mathcal{M}$. A more subtle property is that we can actually draw as many distinct points as we would like from $\mathcal{M}_\epsilon$. We leverage these extra points to build a regularizer which yields good estimations of the local smoothness.
Moreover, taking a discrete approximation of the form in Equation \ref{eq:reg_discrete} with the resampled points from $\mathcal{M}_\epsilon$ still converges to the original operator. Formally, we state the following:

\begin{proposition}
\label{proposition:converge}
    Let $\epsilon, s > 0$ be given. Let $x_1, ..., x_n$ be $n$ samples drawn uniformly at random from a submanifold $\mathcal{M} \subset \mathcal{X}$. For each $x_i$, pick $c$ new points $x_{i,1}, ..., x_{i,c}$ by sampling iid perturbations $\delta_{i,1}, ..., \delta_{i,c}$ and setting $x_{i,j} = x_i + \delta_{i,j}$, where $\forall i,j$, $||\delta_{i,j}|| < \epsilon$. Given a kernel $k_s(x, y) = \exp(-||x - y||^2/s)$, let $L$ and $L_c$ be the Laplacian matrices defined by the $n$ original samples and $n\cdot c$ new samples, respectively. Then if $\epsilon^2 < s$, we have that $L$ and $L_c$ converge to the same operator in the limit as $n \rightarrow \infty$.
\end{proposition}

A proof is sketched in Appendix \ref{appendix:proofs}. For our purposes, this result states that the resampled Laplacian enjoys the same behavior in the limit as the original Laplacian.

\subsection{Sparse approximation of resampled Laplacian}

We next analyze the properties of the Laplacian matrix constructed by the proposed resampling procedure. Although the resampled Laplacian converges to the same operator in the limit, we show that it exhibits behavior we can exploit in the low data regime.

The main observation is that, due to the curse of dimensionality (whereby data becomes increasingly sparse in high dimensions), the resampled Laplacian will consist of two types of edges: very short edges between points resampled from the same $\epsilon$-neighborhood, and very long edges between points sampled from different $\epsilon$-neighborhoods. For an appropriate choice of the scale factor $s$, the exponential form of the kernel causes the weights on the long edges to fall off very quickly compared to the short edges; we thus approximate the Laplacian by dropping long edges and only using weights from short edges. The problem of approximating a graph Laplacian using a subgraph with fewer edges is known as spectral sparsification \citep{spielman2011spectral}. We note that building a sparse approximation $L'$ by sampling edges proportional to their weights yields an unbiased estimator, since:
\begin{align*}
    x^TLx = \sum x_i L_{ij} x_j = \mathbb{E}\bigg[\sum x_i L_{ij} x_j\bigg] = \sum x_i \mathbb{E} [L_{ij}] x_j = x^TL'x
\end{align*}
By the curse of dimensionality, the approximation becomes increasingly accurate as the number of dimensions grows. As a result of our choice of sparsification, the computation of the Laplacian is localized, i.e., the discrete approximation for the intrinsic regularizer can now be written as
\begin{equation}
\label{eq:reg_sparse}
    ||f||^2_I \approx \frac{1}{c^2N^2}\sum_{i = 1}^{N} \sum_{j,k = 1}^c (f(x_{i,j}) - f(x_{i,k}))^2 L(x_{i,j}, x_{i,k})
\end{equation}
The inner loop of this computation involves far fewer terms than the full regularizer (since generally $c <\!\!< n$), and can be efficiently computed pointwise over the full dataset.

We emphasize that the validity of this approximation ultimately depends on the sparsity of the dataset, as well as the particular choice of $\epsilon$ and $s$. In particular, we do not expect this approach to yield better results than the standard manifold regularizer when the data is dense. We provide some rough calculations for CIFAR-10, the setting of our experiments. Each image has dimension 32 x 32 x 3, with pixel values normalized between 0 and 1. The standard deviation for each channel across the dataset is (0.2470, 0.2435, 0.2616). Then even if we assume that each image occupies a ball of radius $0.25$, we would need at least $2^{32*32*3}$ images to cover the entire input space; the 50,000 images in the CIFAR-10 training set cannot even cover 16 dimensions. In comparison, we take $\epsilon = 8/255$, which is about 13\% the standard deviation in each dimension. We conclude that the sparse approximation holds in this setting while the original manifold regularizer without resampling yields almost no information. 

\subsection{Hamming embeddings}

We additionally leverage the structure of ReLU networks to induce a stronger regularization effect on the data manifold. The central observation is that we not only want the function computed by the neural network to be roughly constant on the manifold, we also want the network to produce its outputs in roughly the same way.

First, we identify every local linear component $f_x(\cdot; \theta)$ with its ``activation pattern'', i.e., the sequence of branches taken at each ReLU. Because ReLUs separate the input space via hyperplanes, the activation patterns are unique and induce a Hamming metric on the input space. We call the activation pattern for an input its Hamming embedding, denoted $H(\cdot; \theta)$; similarly, we write $H(\cdot, \cdot; \theta)$ for the Hamming distance between pairs of activations.

Recall that the goal of regularization is to reduce the complexity of the learned function $f$ in the $\epsilon$-neighborhood of inputs $x$. We argue that the Hamming distance provides a concise measure of this complexity. Specifically, if we consider the number of distinct local linear components in the $N_\epsilon(x)$, then $\forall x' \in N_\epsilon(x)$, $H(x, x'; \theta) = 0$ if and only if $N_\epsilon(x)$ is a single linear component. Thus, minimizing the Hamming distance between $x$ and all $x' \in N_\epsilon(x)$ reduces the number of linear components in the neighborhood of $x$. 

\begin{figure}
\centering
    \begin{subfigure}{0.5\textwidth}\centering
        \includegraphics[width=\linewidth]{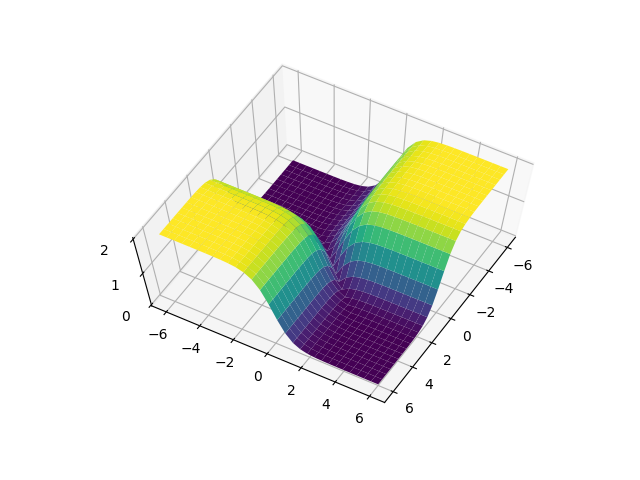} 
    \end{subfigure}
    \begin{subfigure}{0.35\textwidth}\centering
        \includegraphics[width=\linewidth]{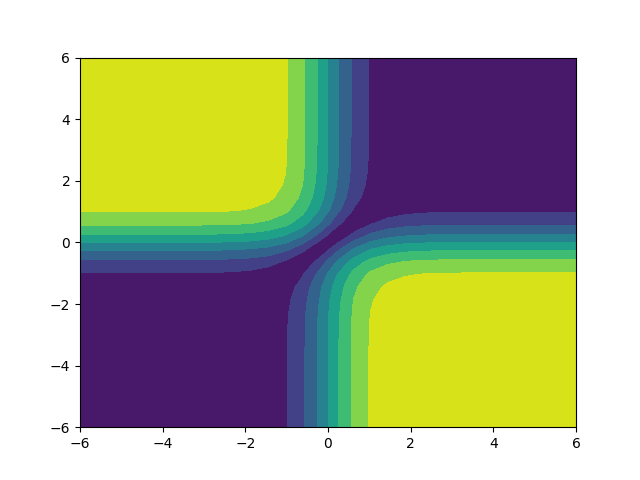}
    \end{subfigure}
\caption{Surface and contour plots for $H_\alpha$ ($\alpha$ = 1).}
\label{HammingPlot}
\end{figure}

Furthermore, we can treat $H(\cdot; \theta)$ as an output of the function $f(\cdot; \theta)$ and write the following manifold regularizer (and its local sparse approximation):
\begin{align*}
    ||H(\cdot; \theta)||^2_I &:= \int_\mathcal{M_\epsilon} ||\nabla_\mathcal{M_\epsilon} H(x; \theta)||^2 d\mu(x) \\
        &\approx \frac{1}{c^2N^2} \sum_{i = 1}^{N} \sum_{j,k = 1}^c H(x_{i,j}, x_{i,k}; \theta)^2 L(x_{i,j}, x_{i,k})
\end{align*}
which is just Equations \ref{eq:reg_integral} and \ref{eq:reg_sparse} with the outputs $f(x)$ replaced by the Hamming embedding. However, this loss term is not continuous in the inputs, and furthermore, the gradients vanish almost everywhere, so it does not generate good training signals. We thus use a continuous relaxation:
\begin{align*}
    H_\alpha(\hat{z}, \hat{y}; \theta) := \text{abs} (\tanh(\alpha * \hat{z}) - \tanh(\alpha * \hat{y}))
\end{align*}
This form is differentiable everywhere except when $\hat{z} = \hat{y}$, and recovers the Hamming distance when we take $\alpha$ to infinity (after scaling). Qualitatively, sensitive activations (i.e., small $|\hat{z}|$ and $|\hat{y}|$) are permitted so long as they are precise. Figure \ref{HammingPlot} presents the surface and contour plots of $H_\alpha$.

Next we observe that, given a set of input points $X$, if $\max_{x, y \in X}H(x, y; \theta) \le c$, then the same bound holds for every point in the convex set of $X$. Thus, it suffices to sample points such that their convex set covers $N_\epsilon(x)$ with high probability. This property motivates generating samples in pairs: randomly sample some corner $x^+$ of $N_\epsilon(x)$ and take the opposing corner $x^- = 2x - x^+$. We can then also drop the weight $L(x_i^+, x_i^-)$ since the distance between $x_i^+$ and $x_i^-$ is constant.

Finally, we note that this idea can be extended more generally to other activation functions by penalizing differing pre-activations more when the second derivative of the activation function is large (and so the first-order Taylor approximation has larger errors).

\subsection{Training with sparse manifold regularization}

For every sample $x$, we generate a new random maximal perturbation $\rho \in \{ \pm \epsilon \}^d$ and produce the pair of perturbed inputs $x^+ := x + p$ and $x^- := x - p$. We compute the standard manifold regularization term as $||f(\cdot, \theta)||^2_I \propto \sum_{i} ||f(x_i^+) - f(x_i^-)||^2_2$. For the Hamming regularizer, we use the $\ell_2$ norm to combine the Hamming distances between pre-activations within a layer and normalize by twice the number of elements in each layer. We sum then normalize by the total number of layers. Quantities are accumulated layer-wise in the forward pass to avoid storing all the preactivations in memory.

The final optimization objective is thus
\begin{align*}
\theta^* = \argmin_\theta \frac{1}{N} \sum_{i = 1}^N V(f(x_i; \theta), y_i) + \gamma_K ||f(\cdot, \theta)||^2_K + \gamma_I ||f(\cdot, \theta)||^2_I  + \gamma_H ||H(\cdot, \theta)||^2_I
\end{align*}
where $V$ is the loss function, $||f(\cdot, \theta)||^2_K$ is the ambient regularizer (e.g., $\ell_2$ loss), and the $\gamma_K, \gamma_I, \gamma_H$ are hyperparameters which control the relative contributions of the different regularizers.

\subsection{Discussion}

Our development has focused on learning deep neural networks which are smooth when samples are sparse over the input manifold $\mathcal{M}$. While this property is related to $\epsilon$-stability, in general the two are not equivalent. Roughly, $\epsilon$-stability is inherently a property about worst case behavior, whereas manifold regularization is aimed at learning a function which is smooth on average. Nonetheless, from a theoretical perspective there are two reasons to expect smoothness to help with stability. The first is that perfect smoothness yields perfect stability (i.e., a constant function on the data manifold); the second is that the two properties converge in the limit as $\epsilon$ approaches zero.

Next, we discuss our choice to focus on the sparse setting. The basic idea is two-fold: first, the requirement that the learned function be $\epsilon$-stable significantly alters the geometry of the input manifold; second, when data is sparse, the amount of information one can glean about the overall geometry of the input manifold is limited, as is evidenced by the vanishing weights in the Laplacian. Our main hypothesis is that the combination of these two observations allows one to formulate a version of the regularizer built from resampled data, which maintains the same properties in the limit but yields more local information when data is sparse.

Conversely, consider the alternative approach, namely, directly learning a smooth function when it is possible to produce samples, not necessarily labelled, that are dense over the input manifold $\mathcal{M}$. Then given the central thesis behind this work, one would expect stability to improve. In fact, the top four results for robust accuracy reported in \citet{croce2020reliable} all exploit an additional dataset of unlabeled images which is orders of magnitude larger than the original training set \citep{carmon2019unlabeled, hendrycks2019using, uesato2019labels, Wang2020Improving}.

\section{Experimental results}
\label{section:results}

\bgroup
\def\arraystretch{1.25}
\begin{table}[]
\caption{CIFAR-10 provable and robust accuracies against an $\ell_\infty$ adversary at $\epsilon = 8/255$.}
\vspace{1em}
\centering
\begin{tabular}{llccc}
    \toprule
    \multicolumn{1}{l}{\multirow{2}{*}{Mechanism}} &
    \multicolumn{1}{l}{\multirow{2}{*}{Defence}} & 
    \multicolumn{3}{c}{Test Accuracy (\%)} \\
     &  & Verified & Robust & Clean \\
	\midrule
	\midrule
	\multirow{2}{*}{Adversarial Training}
	    & PGD (\citet{aleks2017deep}) & & \textbf{45.8} & 87.3 \\
	    & FGSM (\citet{aleks2017deep}) & & 0.0 & 90.3 \\
    \midrule
	\multirow{4}{*}{Certification}
	    & \citet{mirman2018differentiable}$^{\dagger\S}$ & \textbf{35.2} & 40.0 & 54.2 \\
	    & \citet{gowal2018effectiveness} & 32.04 & 34.77 & 49.49 \\
	    & \citet{dvijotham2018training}$^\dagger$ & 26.67 & 32.73 & 48.64 \\
	    & \citet{wong2018scaling} & 21.78 & & 28.67 \\
	\midrule
	\multirow{2}{*}{Verification}
	    & \citet{xiao2019training} & 20.27 & 26.78 & 40.45 \\
	    & \textbf{Manifold Regularization (small)} & 21.04 & 25.56 & 36.66 \\
	\midrule
    \multirow{4}{*}{Regularization}
	    & \citet{pang2019rethinking}$^\ddagger$ & & 24.8 & \textbf{92.7} \\
	    & \textbf{Manifold Regularization (large)} & & 40.54 & 69.95 \\
	    & \hspace{1em} intrinsic regularization only & & 20.11 & 34.74 \\
	    & \hspace{1em} Hamming regularization only & & 24.87 & 90.24 \\
	\bottomrule
	\multicolumn{5}{p{\textwidth - 5em}}{
	    $^\dagger$\footnotesize{With slightly smaller $\epsilon = 0.03 \approx 7.65/255$.} 
	    $^\S$\footnotesize{Computed from 500 / 10000 images in test set.}
	    $^\ddagger$\footnotesize{Results for MMC-10 regularizer only; reported robust accuracy is 55.0\% when trained with PGD.}
	} \\ 
\end{tabular}
\vspace{-1em}
\label{table:robust}
\end{table}
\egroup

We train two models for CIFAR-10 image classification using our regularization techniques: a smaller model for exact verification, following \citet{xiao2019training}; and a larger PreActResNet18 model \citep{he2016identity} for benchmarking against a variety of adversarial models. We also include two ablation studies for the larger model corresponding to using only intrinsic or only Hamming regularization in order to isolate their individual effects on stability. Details about training and model hyperparameters are given in Appendix \ref{appendix:hyperparams}.

The first column of Table \ref{table:robust} reports the \textit{provable robustness} of our trained models and compares against methods which provide similar guarantees, namely by exact verification or certification. This metric is of particular interest for a newly proposed defense such as ours because it establishes a provable lower bound on the robustness achieved by our defense, compared to purely empirical results which may be brittle against more focused attacks \citep{athalye2018obfuscated, carlini2019evaluating, tramer2020adaptive}. Our results reflect state-of-the-art verified robustness against $\ell_\infty$ perturbations of size $\epsilon = 8/255$.

The second column of Table \ref{table:robust} presents robust accuracy against the standard PGD adversary used in the literature. We obtain 40.54\% robust accuracy at $\epsilon = 8/255$ compared to 45.8\% obtained from directly training against the adversary \citep{aleks2017deep}.
For comparison, we also report results for other approaches which are not variants of PGD (i.e., do not rely on an inner maximization problem at each training loop). To the best of our knowledge, \citet{pang2019rethinking} is the only other result in the literature which achieves non-trivial robust accuracy using a regularization only approach. For context, the best result in the literature using a PGD variant is reported by \citet{Wang2020Improving} at 65.04\% robust accuracy. 

Table \ref{table:multipleadv} reports robust accuracy against additional adversaries and perturbation models. For $\ell_2$ and $\ell_\infty$ perturbations against a strong adversary, we use AutoAttack, an ensemble of attacks proposed by \citet{croce2020reliable}. We choose the attack because it is parameter-free, which reduces the possibility of misconfiguration; empirically it is at least as strong as standard PGD. For comparison, we report results from \citet{maini2019adversarial}, who obtain their results after training for 50 epochs with 50 steps of PGD ($\approx$ 2500 effective epochs). We use the results from \citet{aleks2017deep} as a baseline, which show that standard adversarial training against a specific adversary ($\ell_\infty$) does not transfer well to other settings ($\ell_2$). We also report our accuracy against Wasserstein perturbations, which is a metric for images more closely aligned with human perception; we use an implementation of the Wasserstein adversary by \citet{wong2019wasserstein} for our results. They obtain their results using a 50-step inner optimization (number of training epochs is not reported); their baseline model is trained to be certifiably robust against $\ell_\infty$ attacks of size $\epsilon = 2/255$.

With the exception of the exact verification scenario (which uses the same model and setup as in \citet{xiao2019training}), all our results are obtained from a single model after training for 3 hours on one GPU, compared to 80 hours for standard PGD training \citep{shafahi2019adversarial}.

\bgroup
\def\arraystretch{1.25}
\begin{table}[]
\caption{CIFAR-10 robust accuracy against additional adversaries and perturbation models.}
\vspace{1em}
\centering
\begin{tabular}{lcccc}
    \toprule
    \multicolumn{1}{c}{} & \multicolumn{4}{c}{Test Accuracy (\%)} \\
     Defense & $\ell_2$ & $\ell_\infty$ & Wasserstein & Clean \\
     &  $\epsilon = 0.5$ & $\epsilon = 8 / 255$ & $\epsilon = .1$ &  \\
	\midrule
	\midrule
	    \citet{wong2019wasserstein} & & & 76 & 80.69 \\
		\hspace{1em} baseline ($\ell_\infty$ certified) & & & 61 & 66.33 \\
		\midrule
	    \citet{maini2019adversarial} & 66.0 & 49.8$^\ddagger$ & & 81.7 \\
	    \hspace{1em} baseline ($\ell_\infty$ PGD, \citet{aleks2017deep}) & <5.0 & 45.8 & & 87.3 \\
	    \midrule
	    Manifold Regularization & 57.53$^\S$ & 36.90$^\S$ & 66.02 & 69.95 \\
	\bottomrule
	\multicolumn{5}{p{\textwidth - 5em}}{
	    $^\S$\footnotesize{Computed using the full AutoAttack+ attack \citep{croce2020reliable}, which includes APGD-CE, APGD-DLR (+targeted), FAB (+targeted) \citep{croce2019minimally}, and Square Attack \citep{andriushchenko2019square}.}
	    $^\ddagger$\footnotesize{With slightly smaller $\epsilon = 0.03 \approx 7.65/255$.}
	    $^\dagger$\footnotesize{Reported in \citet{maini2019adversarial}.}
	} \\ 
\end{tabular}
\vspace{-1em}
\label{table:multipleadv}
\end{table}
\egroup

\section{Conclusion}

We design regularizers based on manifold regularization that encourage piecewise linear neural networks to learn locally stable functions. We demonstrate this stability by training models to state-of-the-art verified robustness of 21\% against $\ell_\infty$-bounded perturbations of size $\epsilon = 8/255$ on CIFAR-10. We also find that a single model trained using our regularizers is resilient against $\ell_2$, $\ell_\infty$, and Wasserstein-based attacks. 

Critically, computing our regularizers relies only on random sampling, and thus does not require running an inner optimization loop to find strong perturbations at training time. As such, our techniques exhibit strong scaling, since they increase batch sizes rather than epochs during training. This allows us to train our models an order of magnitude faster than standard adversarial training. This work thus presents the first regularization-only approach to achieve comparable results to standard adversarial training against a variety of perturbation models.

\clearpage
\bibliographystyle{plainnat}
\small
\bibliography{ref}
\normalsize

\clearpage
\appendix
\section{Convergence Results}
\label{appendix:proofs}

We first sketch a proof of Proposition \ref{proposition:converge}. Given a kernel $k_s(x, y) = \exp(-||x - y||^2/s)$ and samples $x_1, ..., x_N$, we can define the discrete operator $L$ for an out-of-sample point $x$ as
\begin{align*}
    L f(x) = \frac{1}{N} \sum_{i=1}^N k_s(x, x_i) f(x) - \frac{1}{N} \sum_{i=1}^N k_s(x, x_i) f(x_i)
\end{align*}
We will show pointwise convergence of the original Laplacian $L$ and the resampled Laplacian $L'_c$ to the same continuous operator. Our proof has two parts. First, we show that, for $c = 1$, the resampled Laplacian converges to the original Laplacian. This follows almost immediately from Criterion 5 in \citet{COIFMAN20065}, who use perturbation theory to show that sampling from a manifold with noise yields the same approximation so long as the size of the perturbation $|\delta|$, which in our case is bounded by $\epsilon$, is smaller than the scale parameter of the kernel $\sqrt s$. To see this, note that since our kernel is smooth, we can linearize the effect of the perturbation on the kernel:
\begin{align*}
    k_s(x + \delta_x, y + \delta_y) \le k_s(x, y) + \mathcal{O}\Big(\dfrac{\epsilon}{\sqrt s}\Big)
\end{align*}
Then we also have that the perturbation on $L f(x)$ is of the same order, so that the approximation holds so long as $\epsilon < \sqrt s$, as desired.

Next we show that this holds for arbitrary $c$. This is a simple consequence of the linearity of the operator $L$. Imagine fixing $c$ independently perturbed manifolds and then taking the approximation $L'$ to be the average of the operators defined on each of the perturbed manifolds. Clearly this still converges to the same operator $L$, and since this procedure yields an equivalent formulation of the resampled Laplacian $L'_c$, we are done.

A few additional observations are in order. It turns out that for $L$ to converge to its continuous counterpart in Equation \ref{eq:reg_integral} (i.e., the Laplace-Beltrami operator), we actually need to take the scale of the kernel $s$ to zero (see, e.g. the proofs of convergence in \citet{BELKIN20081289, hein2005graphs, hein2007graph}). Then Proposition \ref{proposition:converge} implies that we also need to take $\epsilon$ to zero, so that in the limit we are just sampling from the unperturbed manifold. In fact, it is possible to prove convergence without taking $s$ to zero, but the continuous operator that is recovered no longer refers to the gradient but rather the value of the kernel over the manifold (\citet{von2008consistency} analyze this case). The spectra of these operators are often used in various clustering algorithms; in the context of deep neural networks, these operators yield similar regularizers for ``smoothness'', though under slightly different definitions.

\clearpage
\section{Experimental Methods and Hyperparameters}
\label{appendix:hyperparams}

We use a PreActResNet18 model \cite{he2016identity} for the CIFAR-10 robustness experiments. We train using SGD and weight decay $\gamma_K$ of 5e-4. We follow \citet{maini2019adversarial} for our learning rate schedule, which is piecewise linear and starts at 0 and goes to 0.1 over first 40 epochs; 0.005 over the next 40 epochs; and 0 over the final 20 epochs. We increase epsilon from 2 to 8 over epochs 10 to 35. We start the weight $\gamma_I$ of the manifold regularization at 0.8 and the weight $\gamma_H$ of the Hamming regularization at 2,400; these increase linearly up to a factor of 10 from epochs 20 to 80. We set the hyperparameter $\alpha = 8$.

We use a two-layer convolutional neural entwork for the CIFAR-10 verification experiments, consisting of 2x2 strided convolutions with 16 and 32 filters, then a 128 hidden unit fully connected layer. This is the same model as used in \citet{wong2018scaling} and \citet{xiao2019training}, except those works use a 100 hidden unit fully connected layer. We use the same schedule for the learning rate and $\epsilon$ as in the PreActResNet18 model. The weight $\gamma_I$ starts at 0.4 and the weight $\gamma_H$ starts at 9000; these increase linearly up to a factor of 10 from epochs 20 to 80. We set the hyperparameter $\alpha = 8$.

For testing, we used a 20-step PGD adversary with 10 random restarts and a step size of $2/255$ as implemented by \citet{robustness}, the full AA+ adversary from \citet{croce2020reliable}, and the Wasserstein adversary from \citet{wong2019wasserstein}. For verification, we adopt the setup of \citet{xiao2019training}, using the MIP verifier of \citet{tjeng2017evaluating}, with solves parallelized over 8 CPU cores and the timeout set to 120 seconds. 

\end{document}